\documentclass{article}

\usepackage[final,nonatbib]{neurips_2022}

\usepackage[utf8]{inputenc} 
\usepackage[T1]{fontenc}    
\usepackage{hyperref}       
\usepackage{url}            
\usepackage{booktabs}       
\usepackage{amsfonts}       
\usepackage{nicefrac}       
\usepackage{microtype}      
\usepackage{xcolor}         
\usepackage{graphicx}
\usepackage{bbm}
\usepackage{amssymb}
\usepackage{amsmath}

\title{An Efficient Two-stage Gradient Boosting Framework for Short-term Traffic State Estimation}

\author{%
  Yichao Lu \\
  Layer 6 AI \\
  \texttt{yichao@layer6.ai} \\
}

\begin{document}

\maketitle

\begin{abstract}
  Real-time traffic state estimation is essential for intelligent transportation systems. The NeurIPS 2022 Traffic4cast challenge provides an excellent testbed for benchmarking short-term traffic state estimation approaches. This technical report describes our solution to this challenge. In particular, we present an efficient two-stage gradient boosting framework for short-term traffic state estimation. The first stage derives the month, day of the week, and time slot index based on the sparse loop counter data, and the second stage predicts the future traffic states based on the sparse loop counter data and the derived month, day of the week, and time slot index. Experimental results demonstrate that our two-stage gradient boosting framework achieves strong empirical performance, achieving third place in both the core and the extended challenges while remaining highly efficient. The source code for this technical report is available at \url{https://github.com/YichaoLu/Traffic4cast2022}.
\end{abstract}

\section{Introduction}

Short-term traffic state estimation is a crucial task in intelligent transportation systems with many practical downstream applications \cite{lv2014traffic,polson2017deep,kreil2020surprising,kopp2021traffic4cast,eichenberger2022traffic4cast}. The NeurIPS 2022 Traffic4cast challenge \footnote{\url{https://www.iarai.ac.at/traffic4cast/challenge/}} provides an excellent testbed for evaluating the performance of short-term traffic state estimation approaches. Given only one hour of sparse loop counter data, the task is to predict the traffic state for all road segments 15 minutes into the future. For the core challenge, the task is to predict the congestion classifications for all road segments. For the extended challenge, the task is to predict the expected time of arrival along super-segments. 

\begin{figure}[!t]
    \centering
    \caption{The relation between average volume and the time slot index for a sample loop counter in London in the Traffic4cast 2022 training data.}
    \includegraphics[width=\textwidth]{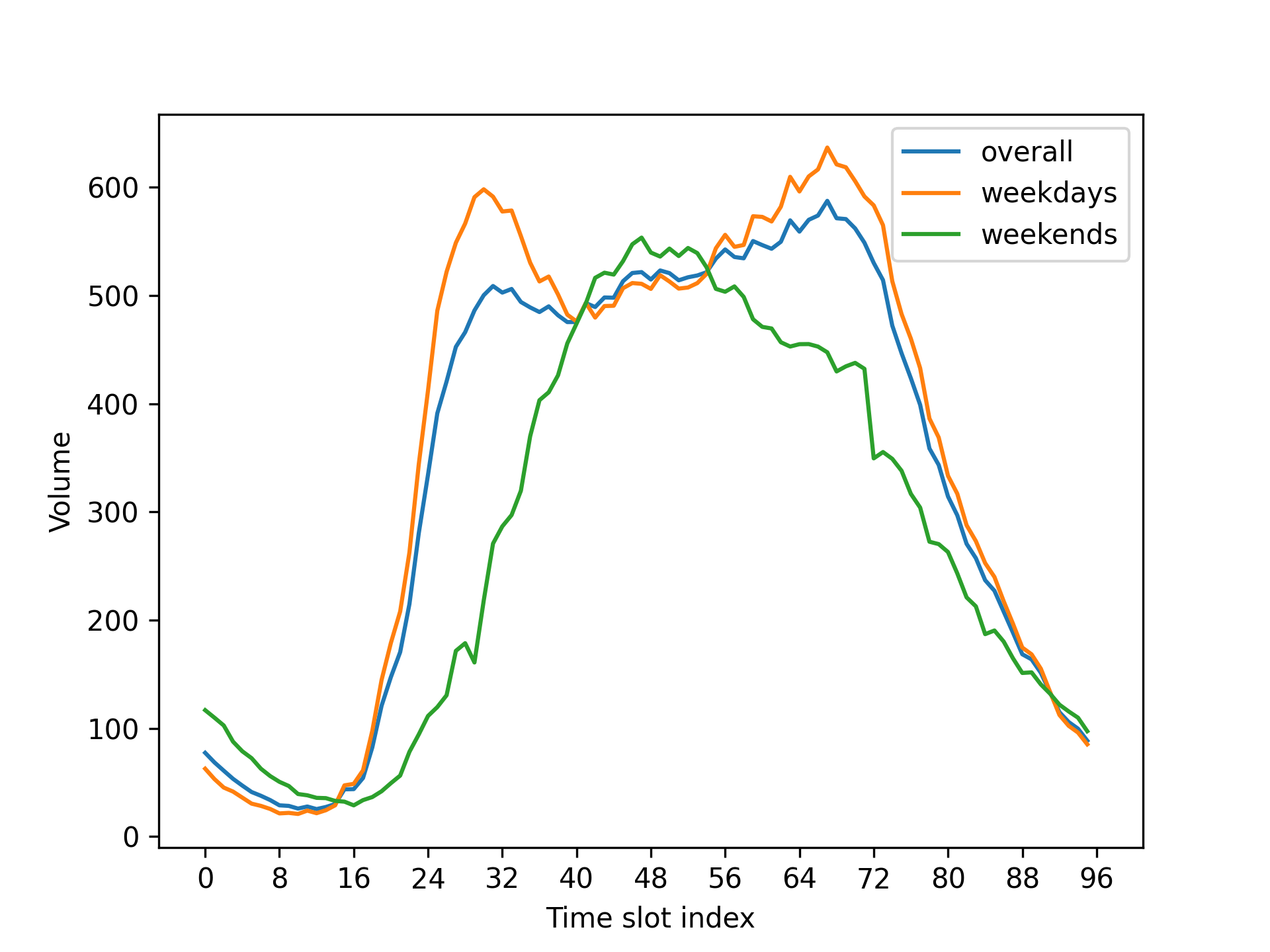}
    \label{fig:seasonality}
\end{figure}

Most state-of-the-art short-term traffic state estimation approaches employ graph neural networks to capture the complex spatio-temporal dependencies among traffic flows \cite{cui2019traffic,wang2020traffic,bui2021spatial,li2021spatial,jiang2022graph}. However, the complexity of urban road networks imposes enormous computational and resource challenges. Therefore, this technical report explores an efficient alternative to graph neural networks for short-term traffic state estimation. In particular, we present a two-stage gradient boosting framework. The advantage of gradient boosting decision trees over neural networks is its high efficiency and the ability to handle missing values without imputation \cite{volkovs2018two,volkovs2019robust,lu2019learning,lu2021multi,lu2021predicting,lu2022session}. The first stage derives the month, day of the week, and time slot index based on the sparse loop counter data, and the second stage predicts the future traffic states based on the sparse loop counter data and the derived month, day of the week, and time slot index. The intuition behind the two-stage approach is as follows. We observe strong seasonality and time trends in traffic flows \cite{li2015trend,yin2016forecasting}. Figure \ref{fig:seasonality} shows the relation between average volume and the time slot index for a sample loop counter in London in the Traffic4cast 2022 training data. Dissecting the task of traffic state estimation into two stages not only harnesses the information within auxiliary labels but also facilitates the second stage model to capture the time patterns in traffic flows.

Experimental results on the NeurIPS 2022 Traffic4cast challenge leaderboards demonstrate that our two-stage gradient boosting framework achieves strong empirical performance, achieving third place in both the core and the extended challenges while remaining highly efficient. The whole pipeline can be trained in less than 3 hours on a single NVIDIA GeForce RTX 2080 Super Mobile GPU.

\section{Approach}

\begin{figure}[!t]
    \centering
    \caption{The relation between average volume and the time slot index for a sample loop counter in London in the Traffic4cast 2022 training data.}
    \includegraphics[width=\textwidth]{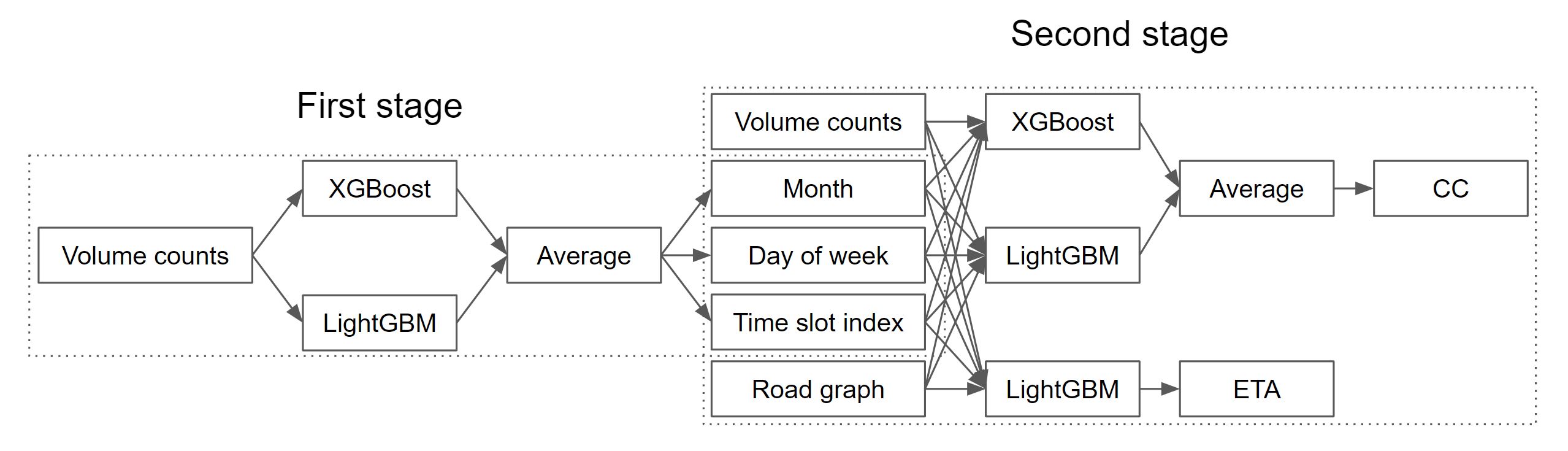}
    \label{fig:architecture}
\end{figure}

The presented two-stage gradient boosting framework is based on two highly efficient implementations of the gradient boosting decision trees algorithm: eXtreme Gradient Boosting (XGBoost) \cite{chen2016xgboost} and Light Gradient Boosting Machine (LightGBM) \cite{ke2017lightgbm}; see Figure \ref{fig:architecture}. The first stage derives the month, day of the week, and time slot index based on the sparse loop counter data. The second stage predicts the future traffic states based on the sparse loop counter data, the road graph attributes, and the predictions from the first stage. The advantage of using a two-stage framework is that it explicitly exploits the strong seasonality and time trends observed in traffic flows; see Figure \ref{fig:seasonality}. As such, the second stage model can better capture the time patterns related to traffic congestion and expected time of arrival. 

\subsection{First Stage}

In the first stage, we use the volume counts for the nodes in the entire road graph as features. We predict the month, day of the week, and time slot index based on the sparse loop counter data. Models based on gradient boosting decision trees are particularly suitable for the first stage task since the loop counter data is sparse and there are many missing values. Tree-based methods can handle missing values by using surrogate splits, thus eliminating the need of imputation. In addition, decision trees are generally more robust to outliers.

We use three separate models for the first stage, as gradient boosting decision trees do not directly support learning multiple targets. All three prediction tasks are modelled as a regression problem, where we optimize the model using L2 loss. The final prediction for the first stage is based on an ensemble of XGBoost and LightGBM models trained separately for each city. We average the predictions from XGBoost and LightGBM models and round the averaged prediction to the nearest integer.

\begin{table}[!th]
  \caption{The list of features and their importance scores (core challenge).}
  \label{features_core}
  \centering
  \begin{tabular}{lll}
    \toprule
    \#     & Description     & Score \\
    \midrule
    1 & TE of cc green given the time slot index and whether the date is a weekend & 674.99 \\
2 & TE of cc red given the time slot index & 432.00 \\
3 & TE of cc green given the time slot index & 301.92 \\
4 & The importance of the highway within the road network & 181.96 \\
5 & TE of cc red given the time slot index and whether the date is a weekend & 166.05 \\
6 & TE of cc yellow given the time slot index and whether the date is a weekend & 135.27 \\
7 & TE of cc red given whether the date is a weekend & 134.05 \\
8 & Whether the date is a weekend & 118.69 \\
9 & TE of cc yellow given the time slot index & 114.36 \\
10 & TE of cc yellow given whether the date is a weekend & 64.03 \\
11 & The time slot index & 55.28 \\
12 & Month & 52.22 \\
13 & TE of cc green given the time slot index and day of the week & 46.05 \\
14 & TE of cc green given day of the week & 45.41 \\
15 & TE of cc red given day of the week & 41.98 \\
16 & Day of the week & 41.39 \\
17 & Whether the road graph edge can only be used in one direction by vehicles & 39.22 \\
18 & Numerical mapping of the OSM highway class & 34.11 \\
19 & TE of cc red given the time slot index and day of the week & 33.58 \\
20 & TE of cc red & 31.34 \\
21 & TE of cc green given whether the date is a weekend & 29.81 \\
22 & Whether the road graph edge runs in a tunnel & 27.98 \\
23 & TE of cc yellow given day of the week & 24.75 \\
24 & Edge speed (km per hour) & 21.54 \\
25 & Maximum legal speed limit & 20.31 \\
26 & TE of cc yellow & 19.80 \\
27 & TE of cc green & 18.83 \\
28 & Number of node hops to get to the closes vehicle counter in the graph & 17.44 \\
29 & Number of traffic lanes on the road graph edge & 14.22 \\
30 & Edge length in meters & 13.69 \\
31 & Unique identifier of the source node & 12.98 \\
32 & Unique identifier of the sink node & 12.82 \\
33 & Unique identifier of the road graph edge & 12.27 \\
34 & Number of incoming edges for the sink node & 12.04 \\
35 & Number of outgoing edges for the source node & 11.97 \\
36 & Number of incoming edges for the source node & 11.78 \\
37 & Number of outgoing edges for the sink node & 11.77 \\
38 & TE of cc yellow given the time slot index and day of the week & 10.74 \\
39 & The volume count for the target node 15 minutes in the past & 8.32 \\
40 & The volume count for the source node 15 minutes in the past & 8.24 \\
41 & The volume count for the target node 45 minutes in the past & 8.06 \\
42 & The volume count for the target node 60 minutes in the past & 8.06 \\
43 & The volume count for the source node 60 minutes in the past & 8.02 \\
44 & The volume count for the source node 45 minutes in the past & 7.90 \\
45 & The volume count for the target node 30 minutes in the past & 7.58 \\
46 & The volume count for the source node 30 minutes in the past & 7.55 \\
    \bottomrule
  \end{tabular}
\end{table}

\begin{table}[!th]
  \caption{The list of features and their importance scores (extended challenge).}
  \label{features_extended}
  \centering
  \begin{tabular}{lll}
    \toprule
    \#     & Description     & Score \\
    \midrule
1 & Smoothed TE given the time slot index and day of the week & 228374728.6 \\
2 & Unique identifier of the supersegment & 76275475.3 \\
3 & Smoothed TE given the time slot index & 44789850.5 \\
4 & The time slot index & 7016971.0 \\
5 & TE given the time slot index and day of the week & 5319797.0 \\
6 & Month & 3703889.9 \\
7 & TE given day of the week & 3033482.3 \\
8 & TE given the time slot index & 2704734.8 \\
9 & Number of nodes in the supersegment & 2693478.1 \\
10 & TE for the supersegment & 1884798.9 \\
11 & Day of the week & 1871912.3 \\
12 & TE given whether the date is a weekend & 1080182.0 \\
13 & Whether the date is a weekend & 935208.7 \\
14 & Smoothed TE given the time slot index and whether the date is a weekend & 804695.4 \\
15 & TE given the time slot index and whether the date is a weekend & 510518.6 \\
    \bottomrule
  \end{tabular}
\end{table}

\subsection{Second Stage}

In the second stage, we apply different models for each challenge since the applicability of gradient boosting decision trees in multi-task learning is limited as opposed to neural networks \cite{chapelle2010multi,chapelle2011boosted,lu2018like,lu2019learning}. At the core of the second stage is extensive feature engineering primarily based on target enconding. Target encoding (TE) calculates the conditional probabilities of the targets given sets of categorical features, which has been demonstrated to be effective in a wide range of machine learning tasks \cite{lu2016context,volkovs2018two,volkovs2019robust,lu2019learning,schifferer2020gpu,deotte2021gpu,lu2021multi,lu2021predicting,lu2021learning,lu2022session,krenn2022predicting}.

\subsubsection{Core Challenge}

For the core challenge, we engineered a number of features capturing the road network characteristics and traffic dynamics. The list of features we have used in the core challenge and their importance scores are presented in Table \ref{features_core}. We can see that the target encodings (TE) of different congestion classes (cc) are among the most important features. For the core challenge, target encoding refers to the fraction of each congestion class in the training set. We apply Bayesian smoothing to reduce overfitting, where the empirical means are the fraction of each congestion class for all road graph edges in the training set; see Equation \ref{eq:te_cc}. 

\begin{equation}
    TE_{cc}(\left[Categories\right]) = \frac{Count(\left[Categories\right]) * mean_{cc}(\left[Categories\right]) + w * mean_{cc}}{Count(\left[Categories\right]) + w}
\label{eq:te_cc}
\end{equation}

In Equation \ref{eq:te_cc}, $Count(\left[Categories\right]$ refers to the number of observations for $Categories$, $mean_{cc}$ is the empirical mean for the congestion class $cc$, and the pseudocount $w > 0$ is a smoothing parameter which we set to $20$.

We train XGBoost and LightGBM using masked cross-entropy loss on congestion classes, which is the same as metric used in the challenge:

\begin{equation}
    \ell(\hat{y}, y) = \sum_{n=1}^N \frac{1}{\Sigma_{n=1}^N   w_{y_n} \cdot \mathbbm{1}\{y_n \not= \text{ignore}\_\text{index}\}} l_n,
\label{eq:loss_cc_1}
\end{equation}
where
\begin{equation}
l_n = - w_{y_n} \log \frac{\hat{y}_{n,y_n}+\varepsilon}{\Sigma_{c=0}^{C-1} \hat{y}_{n,c}+\varepsilon} \cdot \mathbbm{1}\{y_n \not= \text{ignore}\_\text{index}\}
\label{eq:loss_cc_2}
\end{equation}

In Equation \ref{eq:loss_cc_1} and Equation \ref{eq:loss_cc_2}, $\hat{y} \in \mathbbm{R}^{N \times C}$ is the predicted probabilities for each congestion class, $y \in \{0,...,C-1\}^N$ is the target, and $w_c = \frac{N}{|C| \cdot \Sigma_{i=1}^N{   \mathbbm{1} \{y_i = c\} }}$ refers to the class weight that is obtained by calculating the empirical mean for each congestion class. $C \in \mathbbm{N}$ is the number of congestion classes, $N \in \mathbbm{N}$ is the number of samples, $\varepsilon \in \mathbbm{R}^+$ is a small constant to prevent overflow, and $\text{ignore}\_\text{index} \in C \cup \{\bot\}$ specifies whether a target value is ignored or not.

The final prediction for the core challenge is made by an ensemble of XGBoost and LightGBM models, trained separately for each city.

\subsubsection{Extended Challenge}

For the extended challenge, we only use LightGBM since XGBoost does not support the direct optimization of the L1 loss (as of v1.6.2). Unlike conventional gradient boosting methods that work as gradient descent in function space, XGBoost works as Newton-Raphson in function space, and uses a second order Taylor approximation in the loss function to connect to Newton Raphson method \cite{chen2016xgboost}. XGBoost therefore requires smooth objectives to compute the first- and second-order derivative statistics of the loss. On the other hand, L1 loss, also known as mean absolute error (MAE), is not continuously twice differentiable. Thus the direct optimization of the MAE metric is not possible in XGBoost. We experimented with approximating the L1 loss with Huber loss in XGBoost, but could not achieve good performance.

The list of features we have used in the extended challenge and their importance scores are presented in Table \ref{features_extended}. For the extended challenge, target encoding (TE) refers to the average Expected Times of Arrival (ETA) for each supersegment:

\begin{equation}
    TE_{eta}(\left[Categories\right]) = \frac{1}{N}\sum_{i=1}^{N}ETA_{i}(\left[Categories\right])
\end{equation}

We also use smoothed TE, where we use a weighted average of ETAs in the nearby time window to reduce overfitting:

\begin{equation}
    Smoothed\_TE_{eta}(t) = \frac{TE_{eta}(t) + \sum_{i=1}^{4}\left[TE_{eta}(t-i)+TE_{eta}(t+i)\right] * (i+1)^{4}}{1 + \sum_{i=1}^{4}(i+1)^{4}*2}
\end{equation}

The intuition behind smoothed TE is that, the ETAs should be similar within a short time window for any super-segment. Smoothing the target encoding of the ETAs can help prevent overfitting due to outliers. The feature importance scores in Table \ref{features_extended} also demonstrate that smoothed TE is more effective than the non-smoothed counterpart.

For the extended challenge, the task is to predict the ETAs along super-segments. The dynamic speed data obtained from GPS probes is used to derive travel times on the edges of the graph, which are then summed up to derive super-segment ETAs. We experimented with predicting the travel times on the edges of the graph, and then manually summing up the predicted travel times for all edges in the super-segment to estimate super-segment ETA. This results in degraded performance, presumably because using super-segments helps make the ETAs derived from the underlying speed data more robust to data outliers.

\subsection{Implementation Details}

For each city, we randomly select two weeks as the validation set to mimic the distribution of the test set. During training, we use the leave-one-out strategy for target encoding, where the conditional probabilities of the targets are calculated ignoring the targets in the same day to prevent target leakage. During inference, we use all available target values for target encoding.

We use the same hyperparameters in both stages. For XGBoost, we set \textit{max\_depth} to be 5, \textit{eta} (learning rate) to be 0.01, which limits the growth of the trees during training. We set \textit{subsample} to be 0.5, \textit{colsample\_bytree} to be 0.9, and \textit{colsample\_bylevel} to be 0.9 to reduce the risk of overfitting. In addition, \textit{tree\_method} is set to be \textit{gpu\_hist} to use GPU acceleration. The rest of the hyperparameters are set to the default values for XGBoost. For LightGBM, we set \textit{learning\_rate} to be 0.1 and use the default values for the rest of the hyperparameters.

For both XGBoost and LightGBM, we train until the validation score does not improve for 1,000 rounds. This helps estimate the number of rounds that XGBoost and LightGBM require to achieve optimal performance on the held-out set. After that, we retrain the models using both the training set and the validation set for the estimated number of rounds to achieve the optimal performance.

\section{Experiments}

\begin{table}
  \caption{Leaderboard results for the core challenge and the extended challenge.}
  \label{leaderboard}
  \centering
  \begin{tabular}{cllcll}
    \toprule
    \multicolumn{3}{c}{Core challenge} & \multicolumn{3}{c}{Extended challenge}  \\
    \midrule
    Rank     & Team     & Score & Rank     & Team     & Score \\
    \midrule
    1 & ustc-gobbler & 0.84310793876648 & 1 & ustc-gobbler & 58.4997215271 \\
    2 & Bolt & 0.84966790676117 & 2 & TSE & 59.782447814941 \\
    \textbf{3} & \textbf{oahciy (ours)} & \textbf{0.85041532913844} & \textbf{3} & \textbf{oahciy (ours)} & \textbf{61.22274017334} \\
    4 & GongLab & 0.85603092114131 & 4 & Bolt & 61.254610697428 \\
    5 & AP\_DE & 0.87350843350093 & 5 & discovery & 62.296744028727 \\
    \bottomrule
  \end{tabular}
\end{table}


The final leaderboard results on the core and the extended challenges are presented in Table \ref{leaderboard}. Our two-stage gradient boosting framework achieves third place in both challenges, demonstrating its excellent accuracy, generalizability, and transferability.

We further examine the performance of our two-stage gradient boosting framework by conducting a comparative analysis against three alternative methods, namely the multilayer perceptron (MLP), graph neural network (GNN), and the single-stage alternative. We report the leaderboard performance and the training time for each approach in Table \ref{ablation}. We observe that approaches based on gradient boosting decision trees are much more efficient than neural networks based approaches and, at the same time, achieve comparable performance. GNN can outperform MLP while at the cost of much longer training time. Comparing the two-stage framework against the single-stage alternative, we can see that the two-stage framework significantly improves the performance and is only marginally slower than the single-stage model. This is because the first stage model is relatively lightweight and introduces little computational overhead.

To study the effect of the first stage errors on the second stage performance, we comapre the validation performance of the second stage model using the predictions from the first stage model against the same model using the ground truth date and time; see Table \ref{study}. We can see that when the ground truth date and time information is given, the second stage model can achieve significantly better performance, since it prevents errors from being propagated from the first stage to the second stage.

\section{Discussion}

In this challenge, we use the two-stage pipeline because the time of the entries in the test set is not provided as a feature. In a real-world production system, however, the date and the time for the prediction of interest are very easy to obtain. Therefore real-world production systems should directly use the second stage model to avoid the propagation of errors from the first stage to the second stage.

Due to the limit of computational resources, in this challenge, we did not use (graph) neural network-based approaches. Graph neural networks (GNNs) have emerged as a promising short-term traffic state estimation technique due to their ability to model complex spatial dependencies and dynamics \cite{li2021spatial}. GNNs can capture the non-linear interactions between road segments and effectively handle spatio-temporal data, making them suitable for short-term traffic state estimation tasks \cite{bui2022spatial}. The effectiveness of our engineered features when using XGBoost and LightGBM demonstrates that they successfully capture traffic dynamics and time patterns. To further boost the performance of traffic state estimation, we can try adding these engineered features into a deep learning-based framework and ensemble the prediction from deep learning models with the prediction from the gradient boosting approaches.

In our two-stage gradient boosting framework, the first and second stage models are trained separately. We have observed that the errors made by the first stage models can significantly impact the performance of the second stage models. Meanwhile, the predictions from the second stage models (the congestion class and the expected time of arrival) have a very strong dependence on the predictions from the first stage models (the month, day of the week, and time slot index). It is, therefore, possible to apply co-training strategies \cite{blum1998combining,ning2021review} or coevolutionary algorithms \cite{potter1995evolving,garcia2003covnet,lu2018coevolutionary} during the training of the first stage and second stage models, where the predictions from both stages are iteratively refined in a multi-phase fahsion.

\begin{table}
  \caption{Leaderboard results and training time for different approahces.}
  \label{ablation}
  \centering
  \begin{tabular}{cccccc}
    \toprule
    \multicolumn{3}{c}{Core challenge} & \multicolumn{3}{c}{Extended challenge}  \\
    \midrule
    Approach     & Score & Time (minutes) & Approach     & Score & Time (minutes) \\
    \midrule
    MLP & 0.85685 & 197 & MLP & 61.39940 & 172 \\
    GNN & 0.85204 & 1174 & GNN & 61.24305 & 679 \\
    Single-stage & 0.85483 & \textbf{95} & Single-stage & 61.31830 & \textbf{28} \\
    Two-stage & \textbf{0.85041} & 114 & Two-stage & \textbf{61.22274} & 47 \\
    \bottomrule
  \end{tabular}
\end{table}

\section{Related Work}

Short-term traffic state estimation is essential in transportation management as it provides valuable insights into traffic flow patterns and enables better decision-making for commuters and transportation agencies. In recent years, there has been an increasing interest in short-term traffic state estimation due to its potential to improve traffic safety, reduce travel time, and minimize environmental impacts \cite{boukerche2020machine}. Traditionally, short-term traffic state estimation models have relied on classic statistical learning methods such as autoregressive integrated moving average (ARIMA) \cite{dong2009road}, wavelet transform \cite{huang2006forecasting}, and radial basis function network \cite{yang2010traffic}. However, with the success of deep neural networks in capturing complex dependencies and non-linearities in large datasets, researchers have shifted their attention towards utilizing deep neural networks \cite{lv2014traffic,yi2017deep,yu2019crevnet,kreil2020surprising,yuefficient,chen2020citywide,kopp2021traffic4cast,eichenberger2022traffic4cast}.

To model the dynamics of a complex traffic system, a common approach is to represent it as a sequence of movie frames, where each pixel corresponds to the traffic intensity at a certain block of area, and each frame summarizes a discrete time bin, thus casting traffic forecasting as a video prediction task \cite{yu2019crevnet,kreil2020surprising,yuefficient,kopp2021traffic4cast,eichenberger2022traffic4cast}. Early works focused on using convolutional neural networks (CNNs), which are primarily used for visual tasks such as image classification and object detection. To model temporal relationships, a common practice is to concatenate the sequence of frames along the channel dimension and apply 2-dimensional CNNs \cite{yu2019st}. Another approach uses 3-dimensional CNNs, where convolution and pooling operations are performed spatio-temporally \cite{zhang2020network,bayoudh2021transfer}.

Researchers have also explored the potential of recurrent neural networks (RNNs) to capture temporal dependencies \cite{ulbricht1994multi}. However, vanilla RNNs suffer from gradient vanishing and exploding problems, and popular variants such as the Long Short-Term Memory (LSTM) \cite{hochreiter1997long} and the Gated Recurrent Unit (GRU) \cite{chung2014empirical} are often used. State-of-the-art traffic forecasting models have also employed a hybrid of CNN and RNN layers as the underlying architecture, allowing the model to simultaneously exploit the ability of CNN units to model spatial relationships and the potential of RNN units to capture temporal dependencies \cite{shi2015convolutional,shi2017deep,li2017diffusion,yu2019crevnet,yuefficient}.

While CNNs and RNNs are effective in modeling data with an underlying Euclidean or grid-like structure, they fail to capture the complex graph structures in transportation systems such as the road network \cite{kipf2016semi,diehl2019graph,wang2020traffic}. To address this limitation, graph neural networks (GNNs) have recently shown exceptional performance in a variety of traffic flow prediction tasks due to their ability to capture spatial dependencies presented in the form of non-Euclidean graph structures \cite{chen2019gated,lu2021predicting,li2021spatial,jiang2022graph,krenn2022predicting}. Additionally, it has been demonstrated that GNNs can effectively learn properties given by the underlying road network, which improves the generalization performance when making predictions on previously unseen cities \cite{martin2020graph}.

Finally, Transformer-based models have emerged as a powerful tool for short-term traffic state estimation \cite{cai2020traffic,xu2020spatial,yan2021learning}. These models are designed to process sequential data, making them a natural fit for time-series forecasting problems \cite{lu2021context,lu2022session}. One of the key advantages of Transformer-based models is their ability to model long-range dependencies between time steps. This is achieved through the use of self-attention, which allows the model to focus on relevant parts of the input sequence while ignoring irrelevant information \cite{vaswani2017attention}. 

\section{Conclusion}

We present an efficient two-stage gradient boosting framework for short-term traffic state estimation. Experimental results on the NeurIPS 2022 Traffic4cast challenge demonstrate that our two-stage gradient boosting approach achieves excellent accuracy, generalizability, transferability, and computation efficiency.


\begin{table}
  \caption{The effect of the first stage errors on the second stage performance.}
  \label{study}
  \centering
  \begin{tabular}{ccccc}
    \toprule
     & \multicolumn{2}{c}{Core challenge} & \multicolumn{2}{c}{Extended challenge} \\
    \midrule
    & Score & $\Delta$ & Score & $\Delta$ \\
    \midrule
    Prediction & 0.853078 & - & 61.24775 & - \\
    Ground truth & 0.842203 & 1.27\% & 58.03751 & 5.24\% \\
    \bottomrule
  \end{tabular}
\end{table}

\bibliographystyle{abbrv}
\bibliography{main}
\end{document}